\documentclass[conference,hidelinks]{IEEEtran}
\IEEEoverridecommandlockouts
\usepackage{cite}
\usepackage{amsmath,amssymb,amsfonts}
\usepackage{graphicx}
\usepackage{textcomp}
\usepackage{xcolor}
\usepackage{bm}
\usepackage{makecell}
\usepackage{multirow}
\usepackage{threeparttable}
\usepackage{booktabs}
\usepackage{float}
\usepackage{algorithm}
\usepackage{algpseudocode}
\usepackage{hyperref}
\usepackage{orcidlink}
\def\BibTeX{{\rm B\kern-.05em{\sc i\kern-.025em b}\kern-.08em
    T\kern-.1667em\lower.7ex\hbox{E}\kern-.125emX}}
\begin{document}
\title{Titans-QFWP: A Regime-Aware Hybrid Quantum Fast Weight Programmer for Portfolio Optimization\thanks{The views expressed in this article are those of the authors and do not represent the views of Wells Fargo. This article is for informational purposes only. Nothing contained in this article should be construed as investment advice. Wells Fargo makes no express or implied warranties and expressly disclaims all legal, tax, and accounting implications related to this article.}}
\author{
  \IEEEauthorblockN{
    \orcidlink{0009-0002-5141-7909} Ming-Kai Hung\textsuperscript{\dag}
  }
  \IEEEauthorblockA{
    \textit{Dept. of Technology Application} \\
    \textit{and Human Resource Development} \\
    National Taiwan Normal University \\
    Taipei, Taiwan
  }
  \and
  \IEEEauthorblockN{
    \orcidlink{0009-0000-3296-1824} Jun-Hao Chen\textsuperscript{\dag}
  }
  \IEEEauthorblockA{
    \textit{Dept. of Bioenvironmental} \\
    \textit{Systems Engineering} \\
    National Taiwan University \\
    Taipei, Taiwan
  }
  \and
  \IEEEauthorblockN{
    \orcidlink{0000-0002-3830-3272} Yun-Cheng Tsai\textsuperscript{\dag}
  }
  \IEEEauthorblockA{
    \textit{PecuLab LLC} \\
    Seattle, WA, USA
  }
  \and
  \IEEEauthorblockN{
    \orcidlink{0000-0003-0114-4826} Samuel Yen-Chi Chen\textsuperscript{*}
  }
  \IEEEauthorblockA{
    \textit{Wells Fargo} \\
    USA
  }
  \thanks{\textsuperscript{\dag}These authors contributed equally to this work.}
  \thanks{\textsuperscript{*}Corresponding author.}
}
\maketitle
\begin{abstract}
We propose Titans-QFWP, a hybrid reinforcement learning architecture integrating a Quantum Fast Weight Programmer with Titans-style memory (Persistence, Surprise, and Forgetting) for adaptive portfolio optimization. To address high-dimensional market features, we introduce an enhanced A3C\textsuperscript{2} framework with Hungarian-aligned $K$-means clustering and scaled log-return rewards. Evaluated on 468 S\&P~500 stocks under an Equal-Parameter-Count (EPC) benchmark with approximately 3,000 trainable parameters, Titans-QFWP achieves strong performance (median ARR $0.4260$, Calmar $8.5504$, IR $0.8427$). Ablation results reveal that quantum gating fundamentally reshapes memory component roles, with Persistence supporting drawdown control, Surprise contributing to return generation, and Forgetting providing additional stabilization. By stabilizing these quantum representations, the model enables defensive allocation during market drawdowns while preserving upside potential.
\end{abstract}

\section{Introduction}
Financial markets are inherently time-varying \cite{ozbayoglu2020deep} and non-stationary\cite{hamilton1989regime}, driven by regime shifts and secular changes \cite{ang2012regime} that undermine traditional portfolio optimization \cite{fabozzi2008portfolio, lopez2016building}. Deep Reinforcement Learning (DRL) addresses these challenges through sequential policy optimization \cite{cao2025deep, arulkumaran2017deep, henderson2018deep}, but it often struggles with high computational costs \cite{mnih2016asynchronous} and high-dimensional state spaces \cite{yang2020deep}. To manage such expansive spaces, static clustering is typically employed; however, it remains fundamentally incapable of capturing abrupt market transitions \cite{aghabozorgi2015time}. In the quantum domain, Variational Quantum Circuits (VQC) can exploit high-dimensional Hilbert spaces \cite{schuld2019quantum}, yet they lack adaptability due to fixed parameterization \cite{bharti2022noisy, cerezo2021variational, peruzzo2014variational}. Although the Quantum Fast Weight Programmer (QFWP) introduces dynamic weights \cite{chen2024learning, chen2026batched} to overcome this, its update mechanisms often degrade memory stability. To address these challenges, we propose a dynamically adaptive hybrid architecture with two key contributions: 
\begin{enumerate}
\item \textbf{Titans-QFWP Architecture.} Upgrading the original Q-A3C\textsuperscript{2} \cite{liu2026qa3c2}, we replace the VQC with Titans-QFWP. This architecture integrates QFWP with Titans-style memory (Persistence, Surprise, Forgetting) \cite{behrouz2025}. 

\item \textbf{Enhanced A3C\textsuperscript{2} Framework.} We redesign the A3C\textsuperscript{2} reinforcement learning environment \cite{liu2026qa3c2}, incorporating Hungarian-aligned $K$-means clustering, a defensive cash action, and a scaled log-return reward \cite{jiang2017deep}. 
\end{enumerate}
\section{Methodology}
\begin{figure*}[!t]
  \centering
  \includegraphics[width=\textwidth,keepaspectratio]{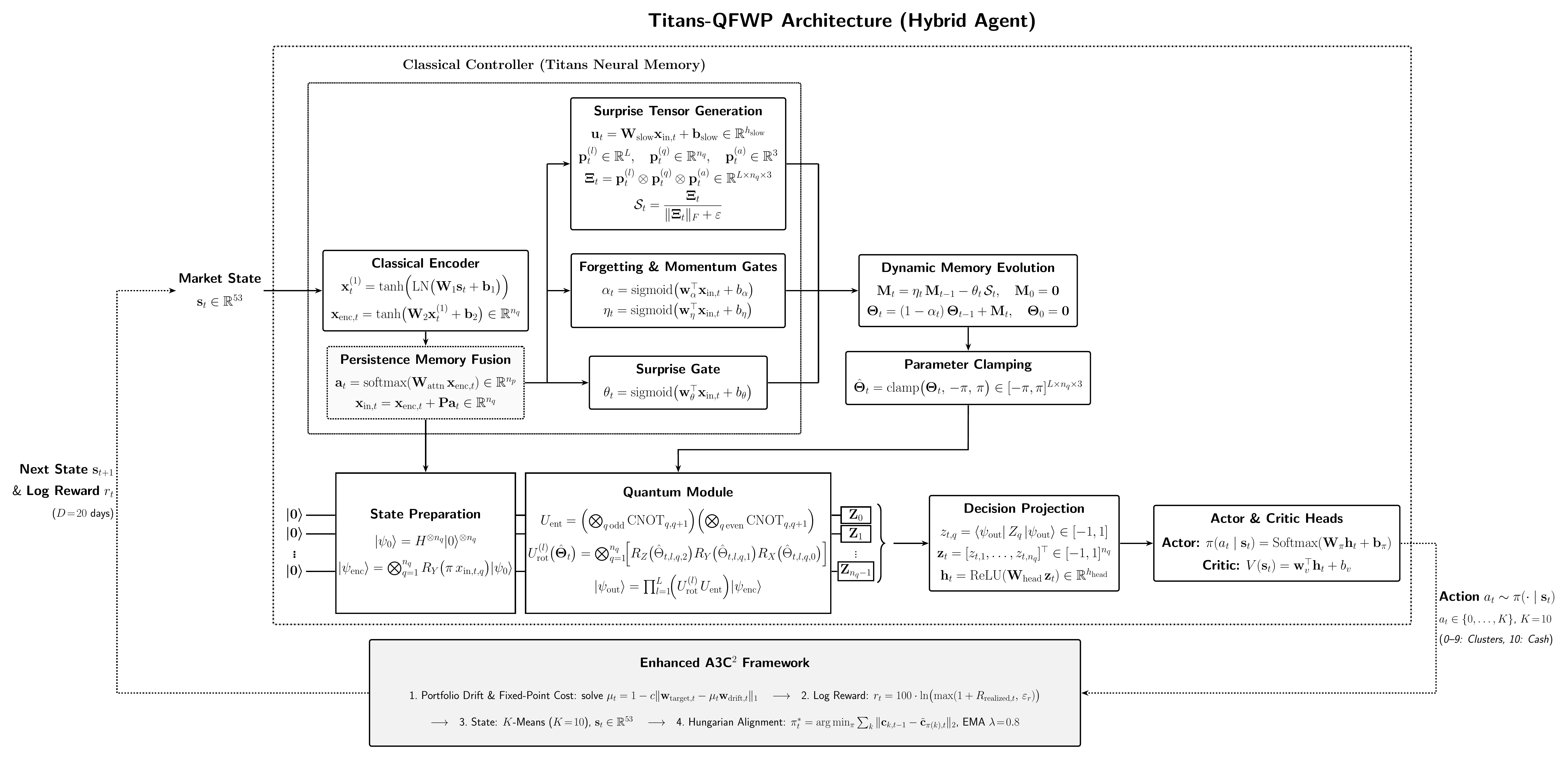}
  \caption{System architecture of the hybrid Titans-QFWP agent integrated with the Enhanced A3C\textsuperscript{2} continuous-reward portfolio simulation environment.}
  \label{fig:architecture}
  \vskip -0.2in
\end{figure*}
\subsection{State Representation and Temporal Alignment}
Using a 60-day lookback, we rebalance every $D=20$ market days. For each asset $i$ at time $t$, the 5- and 20-day moving averages and 20-day volatility of log-returns ($R_{i,t}^{(5)}, R_{i,t}^{(20)}, V_{i,t}^{(20)}$) are cross-sectionally $z$-scored to form $\mathbf{F}_t \in \mathbb{R}^{N \times 3}$ ($N=468$), and then partitioned into $K=10$ clusters via $K$-means. To maintain temporal consistency, the Hungarian algorithm aligns centroids by minimizing $L_2$ drift:
\begin{equation}
  \pi_t^* = \arg\min_{\pi} \sum_{k=0}^{K-1} \|\mathbf{c}_{k,t-1} - \tilde{\mathbf{c}}_{\pi(k),t}\|_2.
  \label{eq:hungarian}
\end{equation}
To filter short-term market noise for the next time step, the matched centroids are smoothed with EMA ($\lambda=0.8$) via $\mathbf{c}_{k,t} = \lambda \mathbf{c}_{k,t-1} + (1-\lambda)\tilde{\mathbf{c}}_{\pi_t^*(k),t}$, initialized based on the 20-day log-return component $c_{k,0}^{(R^{(20)})}$.
Each cluster $\mathcal{C}_k$ yields five features: means $c_{k,t}^{(X)}$ for $X \in \{R^{(5)}, R^{(20)}, V^{(20)}\}$, relative size $|\mathcal{C}_k|/N$, and momentum dispersion $(|\mathcal{C}_k|^{-1} \sum_{i\in\mathcal{C}_k} (R_{i,t}^{(20)}-c_{k,t}^{(R^{(20)})})^2 )^{1/2}$. Concatenating these $5K$ features with three S\&P~500 index features ($R_{spx,t}^{(5)}, R_{spx,t}^{(20)}, V_{spx,t}^{(20)}$) forms the 53-dimensional state $\mathbf{s}_t$ (empty clusters zero-padded).
\subsection{Portfolio Construction and Reward Function}
At each 20-day step $t$, let $\mathbf{w}_{\mathrm{target}, t-1}$ be the prior target weight vector and $\mathbf{G}_t=\exp(\sum_{\tau=1}^{20} \mathbf{r}_\tau)$ the compound gross return vector derived from daily log-returns $\mathbf{r}_\tau$. The drifted weights are calculated as $\mathbf{w}_{\mathrm{drift}, t} = (\mathbf{w}_{\mathrm{target}, t-1} \odot \mathbf{G}_t) / (\mathbf{w}_{\mathrm{target}, t-1}^\top \mathbf{G}_t)$, where $\odot$ denotes the element-wise multiplication. The new weights are then assigned following an inverse volatility strategy, $\mathbf{w}_{\mathrm{target}, t} = \boldsymbol{\sigma}_t^{-1} / \|\boldsymbol{\sigma}_t^{-1}\|_1$, using the 60-day historical volatility vector $\boldsymbol{\sigma}_t$. If the cash action $a_t=K$ is chosen, $\mathbf{w}_{\mathrm{target}, t} = \mathbf{0}$. Rebalancing to these target weights at a cost rate $c=0.0015$ yields a net survival fraction $\mu_t \in (0,1]$, which is solved by successive substitution: $\mu_t = 1 - c\|\mathbf{w}_{\mathrm{target}, t} - \mu_t\mathbf{w}_{\mathrm{drift}, t}\|_1$.
With the gross portfolio return $R_{\mathrm{portfolio}, t} = \mathbf{w}_{\mathrm{target}, t-1}^\top (\mathbf{G}_t - \mathbf{1})$, the net realized return is given by $R_{\mathrm{realized}, t} = \mu_t(1+R_{\mathrm{portfolio}, t})-1$. To ensure stability, the reward floors this return at $10^{-6}$ and scales it by 100, treated as a hyperparameter:
\begin{equation}
  r_t = 100 \cdot \ln \Bigl(\max\bigl(1+R_{\mathrm{realized}, t}, \, 10^{-6}\bigr)\Bigr).
  \label{eq:reward}
\end{equation}
\subsection{Titans-QFWP Architecture and Optimization}
As illustrated in Fig.~\ref{fig:architecture}, the policy fuses Titans-style memory with the Quantum Fast-Weight Programmer (QFWP). First, we apply two $\tanh$ activations to compress the state $\mathbf{s}_t \in \mathbb{R}^{53}$ into an encoded vector $\mathbf{x}_{\mathrm{enc},t} \in \mathbb{R}^{n_q}$ ($n_q = 8$). This is augmented by a trainable matrix $\mathbf{P} \in \mathbb{R}^{n_q \times n_p}$ ($n_p = 4$) through an attention mechanism. The attention weights are computed as $\mathbf{a}_t = \operatorname{softmax}(\mathbf{W}_{\mathrm{attn}}\mathbf{x}_{\mathrm{enc},t}) \in \mathbb{R}^{n_p}$ with $\mathbf{W}_{\mathrm{attn}} \in \mathbb{R}^{n_p \times n_q}$, yielding the augmented representation:
\begin{equation}
  \mathbf{x}_{\mathrm{in},t} = \mathbf{x}_{\mathrm{enc},t} + \mathbf{P}\mathbf{a}_t.
\end{equation}
A slow-program layer projects $\mathbf{x}_{\mathrm{in},t}$ to $\mathbf{u}_t\in\mathbb{R}^{h_{\mathrm{slow}}}$ ($h_{\mathrm{slow}} = 23$). From $\mathbf{u}_t$, three parallel softmax heads generate layer ($\mathbf{p}_t^{(l)}\in\mathbb{R}^{L}$, $L=2$), qubit ($\mathbf{p}_t^{(q)}\in\mathbb{R}^{n_q}$), and axis ($\mathbf{p}_t^{(a)}\in\mathbb{R}^{3}$ for Pauli $X, Y, Z$) distributions, whose normalized outer product forms the Surprise Tensor $\mathcal{S}_t = \boldsymbol{\Xi}_t/(\|\boldsymbol{\Xi}_t\|_F+\varepsilon)$, where $\boldsymbol{\Xi}_t=\mathbf{p}_t^{(l)}\otimes\mathbf{p}_t^{(q)}\otimes\mathbf{p}_t^{(a)}$ and $\varepsilon=10^{-8}$. Conditioned on $\mathbf{x}_{\mathrm{in},t}$, learned sigmoid gates for memory retention ($1 - \alpha_t$), momentum ($\eta_t$), and surprise theta ($\theta_t$) dictate the evolution of fast-weight memory from $\mathbf{M}_0=\boldsymbol{\Theta}_0=\mathbf{0}$:
\begin{equation}
  \mathbf{M}_{t} = \eta_t\mathbf{M}_{t-1} - \theta_t\mathcal{S}_t, \quad \boldsymbol{\Theta}_{t} = (1-\alpha_t)\boldsymbol{\Theta}_{t-1} + \mathbf{M}_{t}.
  \label{eq:memory}
\end{equation}
To serve as valid rotation angles, the raw memory is bounded via $\hat{\boldsymbol{\Theta}}_t = \operatorname{clamp}(\boldsymbol{\Theta}_t, -\pi, \pi) \in [-\pi, \pi]^{L \times n_q \times 3}$ to parameterize the variational quantum circuit. Starting from uniform superposition $|\psi_0\rangle=H^{\otimes n_q}|0\rangle^{\otimes n_q}$, a feature encoding layer $|\psi_{\mathrm{enc}}\rangle = \bigotimes_{q=1}^{n_q} R_Y\bigl(\pi\,x_{\mathrm{in},t,q}\bigr)|\psi_0\rangle$ precedes $L$ layers of CNOT entanglers $U_{\mathrm{ent}}$ and parameterized rotations $U_{\mathrm{rot}}^{(l)}\bigl(\hat{\boldsymbol{\Theta}}_t\bigr) =
\bigotimes_{q=1}^{n_q}\!\Bigl[R_Z\bigl(\hat{\Theta}_{t,l,q,2}\bigr) R_Y\bigl(\hat{\Theta}_{t,l,q,1}\bigr)
R_X\bigl(\hat{\Theta}_{t,l,q,0}\bigr)\Bigr]$. Finally, the output state $|\psi_{\mathrm{out}}\rangle = \prod_{l=1}^{L}(U_{\mathrm{rot}}^{(l)}U_{\mathrm{ent}})|\psi_{\mathrm{enc}}\rangle$ yields the expectation vector $\mathbf{z}_t \in [-1,1]^{n_q}$, composed of the Pauli-Z expectation values $\langle \psi_{\text{out}} | Z_q | \psi_{\text{out}} \rangle$ for each qubit. This vector is subsequently processed through a ReLU activation function to generate actor and critic heads for cluster selection, with the constituent stocks weighted by inverse volatility.
\section{Experimental Setup}
\noindent\textbf{Dataset.} Daily log-returns of S\&P~500 stocks (Jan.~2015--Apr.~2026) are used. Assets with more than 10\% missing observations are excluded, yielding 468 stocks. All features are constructed using only the information available at time $t$. The data are divided into 2,262 training days (2015-01-06 to 2023-12-29), 252 validation days (2024-01-02 to 2024-12-31), and 332 test days (2025-01-02 to 2026-04-30).

\textbf{Equal-Parameter-Count (EPC) Benchmark.} 
For fair comparison, all 11 architectures are limited to approximately 3,000 trainable parameters. Inspired by Titans (Memory-as-Context) \cite{behrouz2025}, our proposed Titans-QFWP (Full) is evaluated alongside its variants (allocations detailed in Table~\ref{tab:parameters}).
\begin{table}[!t]
  \centering
  \caption{Parameter configuration for the EPC benchmark.}
  \label{tab:parameters}
  \footnotesize
  \begin{tabular}{lrrr}
    \toprule
    \textbf{Architecture} & \textbf{Total} & \textbf{Quantum} & \textbf{Classical} \\
    \midrule
    \multicolumn{4}{l}{\textit{Classical Architectures}} \\
    \quad Titans (Memory-as-Context) & 2,997 & 0 & 2,997 \\
    \quad FWP                        & 3,000 & 0 & 3,000 \\
    \quad Titans-FWP (Full)          & 2,999 & 0 & 2,999 \\
    \quad Titans-FWP ($-$Forgetting) & 2,990 & 0 & 2,990 \\
    \quad Titans-FWP ($-$Persistence) & 2,931 & 0 & 2,931 \\
    \quad Titans-FWP ($-$Surprise)   & 2,981 & 0 & 2,981 \\
    \midrule
    \multicolumn{4}{l}{\textit{Quantum Architectures}} \\
    \quad QFWP                       & 3,001 & 261   & 2,740 \\
    \quad Titans-QFWP (Full)         & 3,000 & 1,253 & 1,747 \\
    \quad Titans-QFWP ($-$Forgetting) & 2,991 & 1,253 & 1,738 \\
    \quad Titans-QFWP ($-$Persistence) & 2,932 & 1,253 & 1,679 \\
    \quad Titans-QFWP ($-$Surprise)   & 2,982 & 1,253 & 1,729 \\
    \bottomrule
  \end{tabular}
  \vskip -0.2in
\end{table}
\section{Experimental Results and Discussion}
\subsection{Overall Analysis}
Results over 10 random seeds (1--10), reported as median [IQR] in Table~\ref{tab:oos_performance}, are compared with the market benchmark (S\&P 500 Buy \& Hold). Titans-QFWP (Full) achieves higher Annualized Rate of Return (ARR; $0.4260$ vs. $0.2528$), Calmar Ratio (Calmar; $8.5504$ vs. $7.1638$), and the highest Information Ratio (IR; $0.8427$). In contrast, the market benchmark exhibits lower Maximum Drawdown (MDD; $-0.0353$) and higher Sortino Ratio (Sortino; $8.5405$), partly due to the exclusion of transaction costs. These results demonstrate that Titans-QFWP effectively captures market trends under the strong bull-market regime.
\begin{table*}[!t]
  \centering
  \caption{S\&P 500 out-of-sample performance across asset allocation baselines (median [IQR], seeds 1--10).}
  \label{tab:oos_performance}
  \footnotesize
  \resizebox{\textwidth}{!}{
  \begin{tabular}{lrrrrr} 
    \toprule
    \textbf{Architecture} & \multicolumn{1}{c}{\textbf{ARR}} & \multicolumn{1}{c}{\textbf{MDD}} & \multicolumn{1}{c}{\textbf{Sortino}} & \multicolumn{1}{c}{\textbf{Calmar}} & \multicolumn{1}{c}{\textbf{IR}} \\
    \midrule
    \multicolumn{6}{l}{\textit{Market Benchmark}} \\
    \quad S\&P 500 (Buy \& Hold)
      & $0.2528$ & $-0.0353$ & $8.5405$ & $7.1638$ & \multicolumn{1}{c}{N/A} \\ 
    \midrule
    \multicolumn{6}{l}{\textit{Baseline \& Classical Architectures}} \\
    \quad Titans (Memory-as-Context)
      & $0.2557\;\;[0.2649]$ & $\mathbf{-0.0512}\;\;[\mathbf{0.0355}]$ & $3.3645\;\;[4.4568]$ & $3.9077\;\;[7.2651]$ & $0.1374\;\;[1.8125]$ \\
    \quad FWP
      & $\mathbf{0.3423}\;\;[\mathbf{0.1904}]$ & $-0.0918\;\;[0.1046]$ & $3.3498\;\;[6.1209]$ & $3.9429\;\;[7.7964]$ & $\mathbf{0.4447}\;\;[\mathbf{1.3117}]$ \\
    \quad Titans-FWP (Full)
      & $0.3175\;\;[0.2013]$ & $-0.0606\;\;[0.0648]$ & $\mathbf{4.9382}\;\;[\mathbf{3.6665}]$ & $\mathbf{6.2776}\;\;[\mathbf{5.6673}]$ & $0.3891\;\;[0.9531]$ \\
    \quad Titans-FWP ($-$Forgetting)
      & $0.2752\;\;[0.1713]$ & $-0.0790\;\;[0.0453]$ & $3.5281\;\;[3.4122]$ & $3.4519\;\;[4.0856]$ & $0.1993\;\;[1.0045]$ \\
    \quad Titans-FWP ($-$Persistence)
      & $0.2841\;\;[0.2500]$ & $-0.0702\;\;[0.0540]$ & $4.1309\;\;[2.8475]$ & $4.5385\;\;[3.4245]$ & $0.1927\;\;[1.2757]$ \\
    \quad Titans-FWP ($-$Surprise)
      & $0.1907\;\;[0.1834]$ & $-0.0774\;\;[0.0731]$ & $2.8587\;\;[3.8253]$ & $2.5238\;\;[6.5771]$ & $-0.4547\;\;[1.1375]$ \\
    \midrule
    \multicolumn{6}{l}{\textit{Quantum Architectures}} \\
    \quad QFWP
      & $0.1806\;\;[0.1437]$ & $-0.1017\;\;[0.0887]$ & $1.8849\;\;[1.1011]$ & $1.5299\;\;[2.2969]$ & $-0.4002\;\;[0.8815]$ \\
    \quad \textbf{Titans-QFWP (Full)}
      & $\mathbf{0.4260}\;\;[\mathbf{0.4197}]$ & $\mathbf{-0.0500}\;\;[\mathbf{0.0212}]$ & $\mathbf{6.9018}\;\;[\mathbf{6.5930}]$ & $\mathbf{8.5504}\;\;[\mathbf{6.0697}]$ & $\mathbf{0.8427}\;\;[\mathbf{1.3992}]$ \\
    \quad Titans-QFWP ($-$Forgetting)
      & $0.1965\;\;[0.1276]$ & $-0.0520\;\;[0.0415]$ & $3.0186\;\;[1.3646]$ & $3.3316\;\;[2.0293]$ & $-0.3887\;\;[0.9337]$ \\
    \quad Titans-QFWP ($-$Persistence)
      & $0.1929\;\;[0.1297]$ & $-0.1010\;\;[0.0400]$ & $2.0905\;\;[1.1939]$ & $2.1663\;\;[1.7055]$ & $-0.3136\;\;[0.6757]$ \\
    \quad Titans-QFWP ($-$Surprise)
      & $0.1586\;\;[0.1749]$ & $-0.0782\;\;[0.0318]$ & $2.6949\;\;[1.5092]$ & $2.3728\;\;[1.3064]$ & $-0.5138\;\;[1.1810]$ \\
    \bottomrule
  \end{tabular}
  }
  \vskip -0.15in
\end{table*}

\subsection{Ablation Diagnostics}
\subsubsection{Classical Ablation (Titans-FWP)}
In classical variants (Table~\ref{tab:ablation_fwp}), removing Surprise produces the largest degradation in ARR, Sortino, Calmar, and IR. Removing Forgetting results in the largest deterioration in MDD. Removing Persistence leads to smaller performance changes across most metrics.
\begin{table*}[!t]
  \centering
  \caption{Performance metrics for Titans-FWP ablation variants (median [IQR], seeds 1--10).}
  \label{tab:ablation_fwp}
  \footnotesize
  \resizebox{\textwidth}{!}{
  \begin{tabular}{lrrrrr} 
    \toprule
    \textbf{Architecture} & \multicolumn{1}{c}{\textbf{ARR}} & \multicolumn{1}{c}{\textbf{MDD}} & \multicolumn{1}{c}{\textbf{Sortino}} & \multicolumn{1}{c}{\textbf{Calmar}} & \multicolumn{1}{c}{\textbf{IR}} \\
    \midrule
    \multicolumn{6}{l}{\textit{Classical Ablation Study: Titans-FWP}} \\
    \quad FWP (Baseline)
      & $\mathbf{0.3423}\;\;[\mathbf{0.1904}]$ & $-0.0918\;\;[0.1046]$ & $3.3498\;\;[6.1209]$ & $3.9429\;\;[7.7964]$ & $\mathbf{0.4447}\;\;[\mathbf{1.3117}]$ \\
    \quad \textbf{Titans-FWP (Full)}
      & $0.3175\;\;[0.2013]$ & $\mathbf{-0.0606}\;\;[\mathbf{0.0648}]$ & $\mathbf{4.9382}\;\;[\mathbf{3.6665}]$ & $\mathbf{6.2776}\;\;[\mathbf{5.6673}]$ & $0.3891\;\;[0.9531]$ \\
    \quad Titans-FWP ($-$Forgetting)
      & $0.2752\;\;[0.1713]$ & $-0.0790\;\;[0.0453]$ & $3.5281\;\;[3.4122]$ & $3.4519\;\;[4.0856]$ & $0.1993\;\;[1.0045]$ \\
    \quad Titans-FWP ($-$Persistence)
      & $0.2841\;\;[0.2500]$ & $-0.0702\;\;[0.0540]$ & $4.1309\;\;[2.8475]$ & $4.5385\;\;[3.4245]$ & $0.1927\;\;[1.2757]$ \\
    \quad Titans-FWP ($-$Surprise)
      & $0.1907\;\;[0.1834]$ & $-0.0774\;\;[0.0731]$ & $2.8587\;\;[3.8253]$ & $2.5238\;\;[6.5771]$ & $-0.4547\;\;[1.1375]$ \\
    \bottomrule
  \end{tabular}
  }
  \vskip -0.15in
\end{table*}

\subsubsection{Quantum Ablation (Titans-QFWP)}
In quantum variants (Table~\ref{tab:ablation_qfwp}), removing Persistence causes the largest degradation in the median MDD, Sortino, and Calmar, while removing Surprise leads to the lowest median ARR and IR. These patterns suggest that quantum gating alters memory roles, with greater expressive capacity favoring Persistence over Forgetting for representation preservation.

\begin{table*}[!t]
  \centering
  \caption{Performance metrics for Titans-QFWP ablation variants (median [IQR], seeds 1--10).}
  \label{tab:ablation_qfwp}
  \footnotesize
  \resizebox{\textwidth}{!}{
  \begin{tabular}{lrrrrr} 
    \toprule
    \textbf{Architecture} & \multicolumn{1}{c}{\textbf{ARR}} & \multicolumn{1}{c}{\textbf{MDD}} & \multicolumn{1}{c}{\textbf{Sortino}} & \multicolumn{1}{c}{\textbf{Calmar}} & \multicolumn{1}{c}{\textbf{IR}} \\
    \midrule
    \multicolumn{6}{l}{\textit{Quantum Ablation Study: Titans-QFWP}} \\
    \quad QFWP (Baseline)
      & $0.1806\;\;[0.1437]$ & $-0.1017\;\;[0.0887]$ & $1.8849\;\;[1.1011]$ & $1.5299\;\;[2.2969]$ & $-0.4002\;\;[0.8815]$ \\
    \quad \textbf{Titans-QFWP (Full)}
      & $\mathbf{0.4260}\;\;[\mathbf{0.4197}]$ & $\mathbf{-0.0500}\;\;[\mathbf{0.0212}]$ & $\mathbf{6.9018}\;\;[\mathbf{6.5930}]$ & $\mathbf{8.5504}\;\;[\mathbf{6.0697}]$ & $\mathbf{0.8427}\;\;[\mathbf{1.3992}]$ \\
    \quad Titans-QFWP ($-$Forgetting)
      & $0.1965\;\;[0.1276]$ & $-0.0520\;\;[0.0415]$ & $3.0186\;\;[1.3646]$ & $3.3316\;\;[2.0293]$ & $-0.3887\;\;[0.9337]$ \\
    \quad Titans-QFWP ($-$Persistence)
      & $0.1929\;\;[0.1297]$ & $-0.1010\;\;[0.0400]$ & $2.0905\;\;[1.1939]$ & $2.1663\;\;[1.7055]$ & $-0.3136\;\;[0.6757]$ \\
    \quad Titans-QFWP ($-$Surprise)
      & $0.1586\;\;[0.1749]$ & $-0.0782\;\;[0.0318]$ & $2.6949\;\;[1.5092]$ & $2.3728\;\;[1.3064]$ & $-0.5138\;\;[1.1810]$ \\
    \bottomrule
  \end{tabular}
  }
  \vskip -0.2in
\end{table*}
\subsection{Portfolio Dynamics and Asset Selection}
\begin{figure}[!t]
  \centering
  \includegraphics[width=\columnwidth,height=0.24\textheight,keepaspectratio]{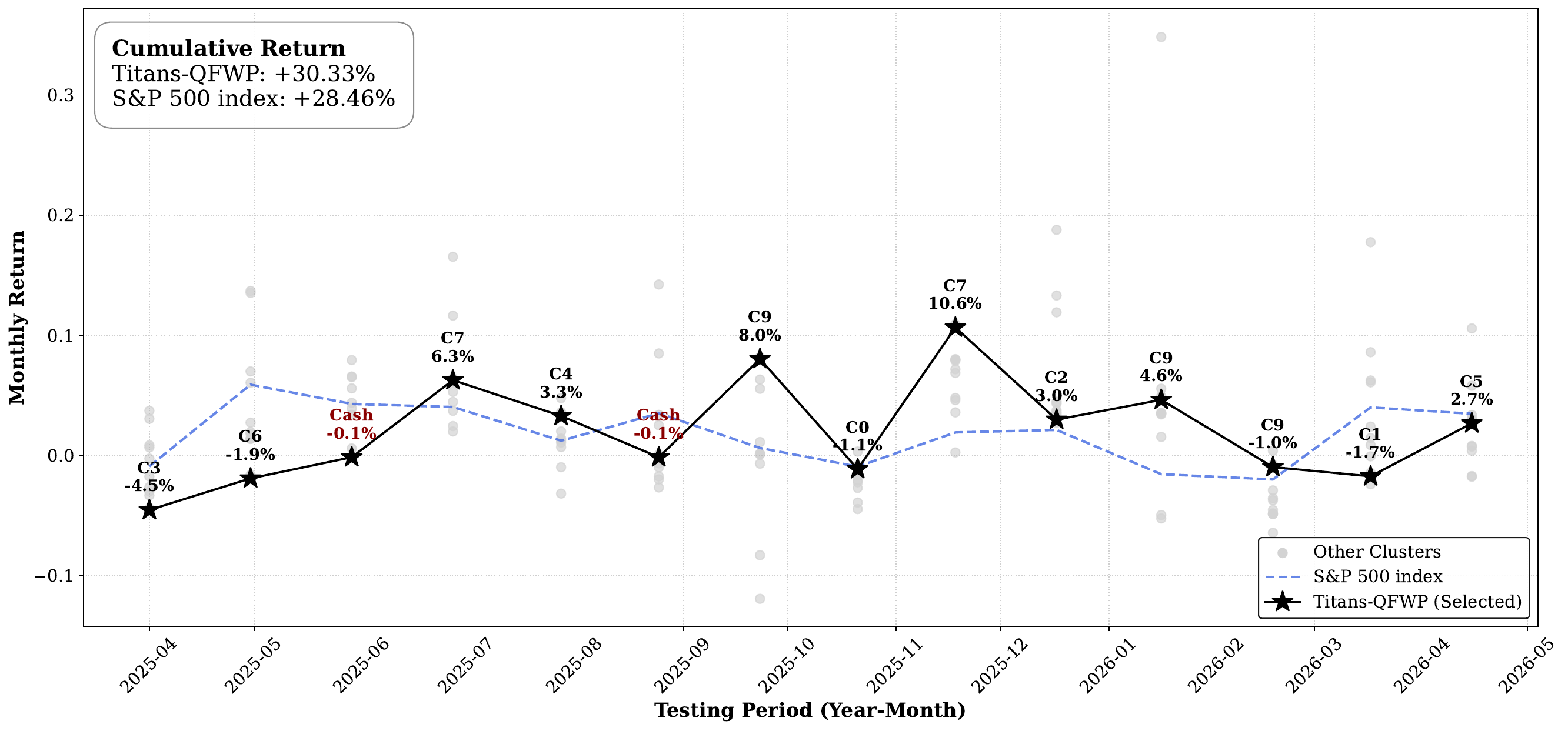}
  \caption{Dynamic cluster selection of Titans-QFWP vs. the S\&P~500 index.}
  \label{fig:selection_analysis}
  \vskip -0.1in
\end{figure}

As shown in Fig.~\ref{fig:selection_analysis} (Seed 1), Titans-QFWP achieves a cumulative return of +30.33\%, slightly outperforming the S\&P 500 (+28.46\%), driven by allocations to higher-return clusters (e.g., Clusters 7 and 9). The model adaptively reallocates to cash during market downturns.

\begin{figure}[!t]
  \centering
  \includegraphics[width=\columnwidth,keepaspectratio]{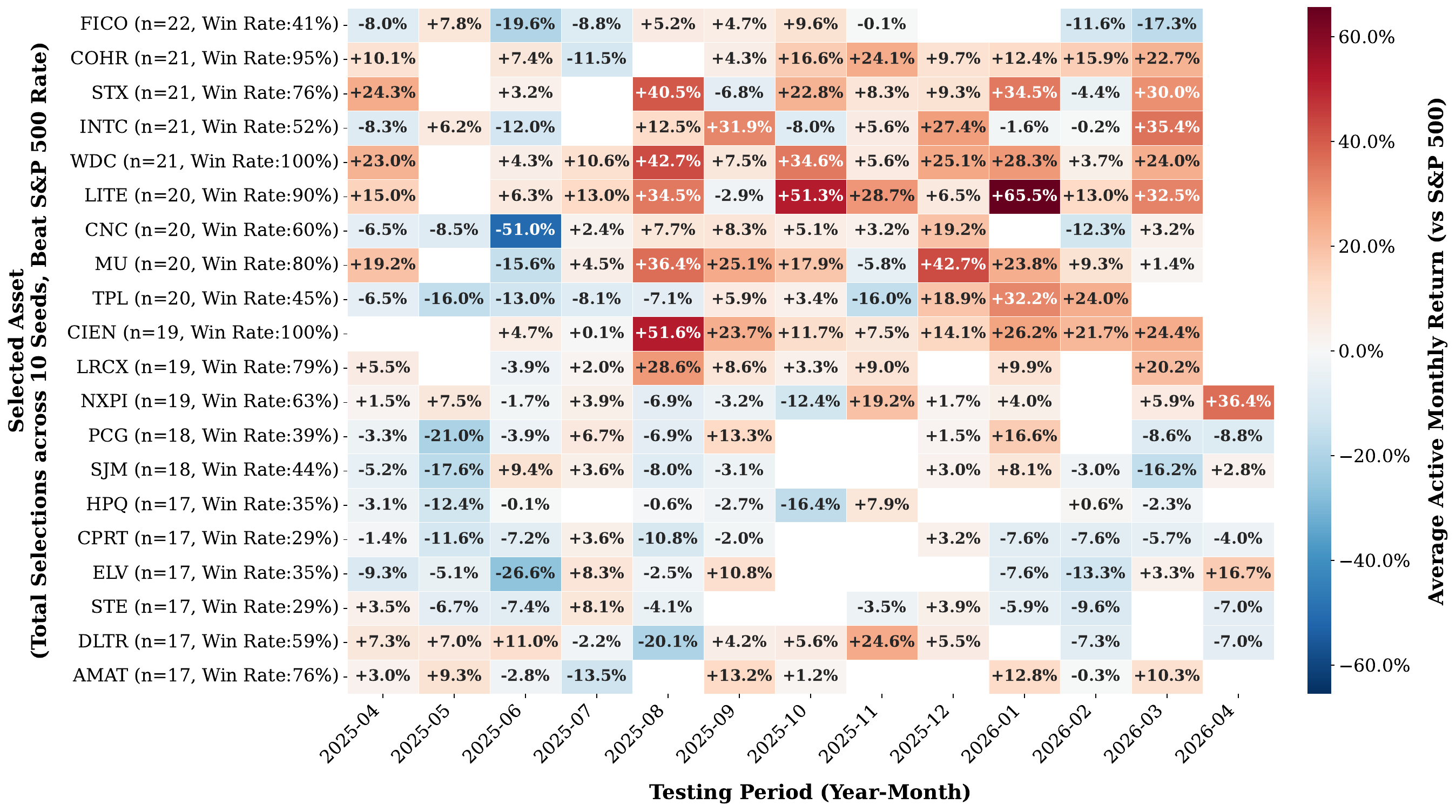}
  \caption{Top-20 selected stocks of Titans-QFWP vs. the S\&P~500 index.}
  \label{fig:heatmap_active_return}
  \vskip -0.2in
\end{figure}

Averaged across 10 seeds (1--10), the top-20 selected stocks exhibit concentrated outperformance in technology sub-sectors, including data storage, optical networking, photonics, and memory semiconductors (Fig.~\ref{fig:heatmap_active_return}). Measured by active return (excess simple return over the S\&P 500), WDC and CIEN outperform the benchmark at every time step, with peak active returns of +42.7\% and +51.6\%. In addition, COHR, LITE, and MU also contribute substantial excess returns.

\section{Limitations}
Limitations include survivorship bias, a short out-of-sample period, seed variability, simplified transaction costs, and evaluation of quantum components using \texttt{lightning.qubit} quantum simulation. Risk-adjusted metrics are reported descriptively rather than as  statistically conclusive evidence.
\section{Conclusion}
Integrating QFWP with Titans-style memory provides a robust framework for addressing market non-stationarity. Crucially, the expressive capacity of quantum gating fundamentally alters learning dynamics: it shifts the mechanism of representation preservation, relying more on Persistence rather than Forgetting for drawdown control. Although frictionless passive benchmarks exhibit a marginal downside advantage during bull markets, Titans-QFWP mitigates this through regime-aware allocation, dynamically transitioning between defensive and high-performing asset clusters. These findings illustrate how quantum-classical memory architectures reorganize representation learning, advancing adaptive financial reinforcement learning.
\bibliographystyle{IEEEtran}
\bibliography{reference}
\end{document}